\documentclass[10pt,twocolumn,letterpaper]{article}

\usepackage{cvpr}              

\usepackage{graphicx}
\usepackage{amsmath}
\usepackage{amssymb}
\usepackage{booktabs}
\usepackage{url}
\usepackage{floatrow}
\usepackage{graphicx}
\graphicspath{ {./images/} }
\usepackage{caption}
\usepackage{subcaption}
\usepackage[accsupp]{axessibility}  

\usepackage[pagebackref,breaklinks,colorlinks]{hyperref}

\usepackage[capitalize]{cleveref}
\crefname{section}{Sec.}{Secs.}
\Crefname{section}{Section}{Sections}
\Crefname{table}{Table}{Tables}
\crefname{table}{Tab.}{Tabs.}

\def\cvprPaperID{20}
\def\confName{CVPR}
\def\confYear{2023}

\begin{document}

\title{Multispectral Contrastive Learning with Viewmaker Networks}

\author{Jasmine Bayrooti, Noah Goodman, Alex Tamkin\\
Department of Computer Science\\
Stanford University\\
Stanford, CA 94305, USA\\
{\tt\small \{jbayrooti, ngoodman, atamkin\}@stanford.edu}
}
\maketitle

\begin{abstract}
   Contrastive learning methods have been applied to a range of domains and modalities by training models to identify similar ``views'' of data points. However, specialized scientific modalities pose a challenge for this paradigm, as identifying good views for each scientific instrument is complex and time-intensive. In this paper, we focus on applying contrastive learning approaches to a variety of remote sensing datasets. We show that Viewmaker networks, a recently proposed method for generating views without extensive domain knowledge, can produce useful views in this setting. We also present a Viewmaker variant called Divmaker, which achieves similar performance and does not require adversarial optimization. Applying both methods to four multispectral imaging problems, each with a different format, we find that Viewmaker and Divmaker can outperform cropping- and reflection-based methods for contrastive learning in every case when evaluated on downstream classification tasks. This provides additional evidence that domain-agnostic methods can empower contrastive learning to scale to real-world scientific domains. Open source code can be found at \url{https://github.com/jbayrooti/divmaker}.
\end{abstract}

\begin{figure*}
  \centering
  \begin{subfigure}{\linewidth}
    \includegraphics[width=\linewidth]{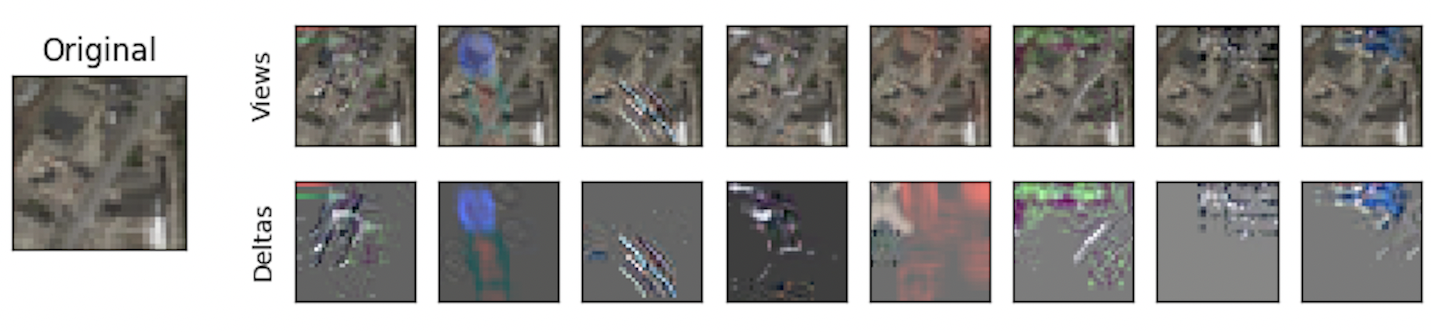}
    \caption{The input image is from NWPU-RESISC45 with the label Baseball Diamond.}
  \end{subfigure}
  \vfill
  \begin{subfigure}{\linewidth}
    \includegraphics[width=\textwidth]{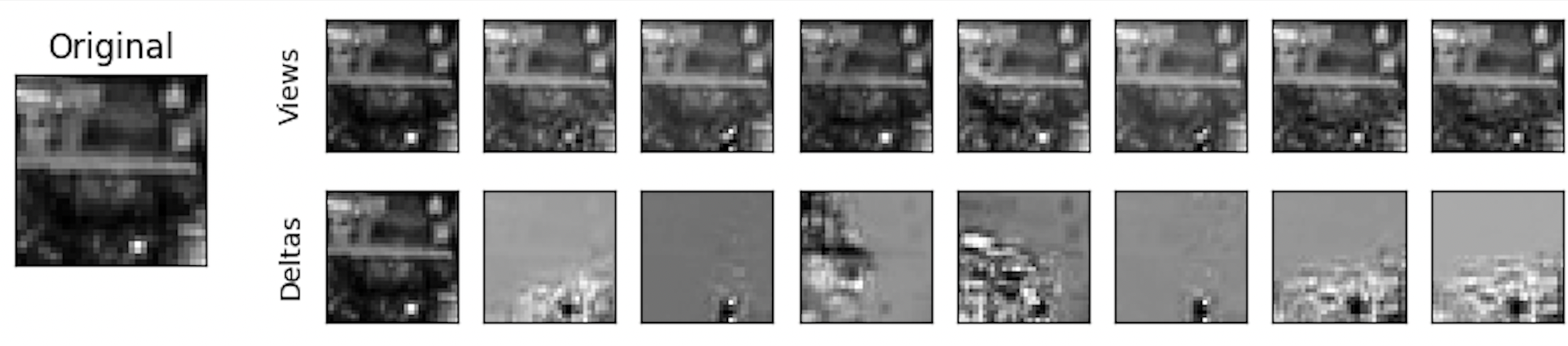}
    \caption{The input image is band 2 from So2Sat Sentinel-2 with the label Compact Low-Rise.}
  \end{subfigure}
  \vfill
  \begin{subfigure}{\linewidth}
    \includegraphics[width=\textwidth]{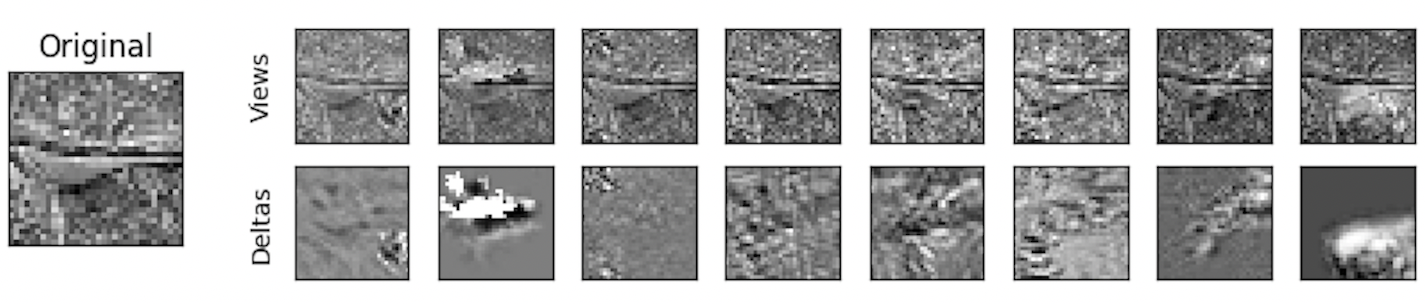}
    \caption{The input image is band 1 from BigEarthNet with the multi-label: Mixed Forest, Transitional Woodland/Shrub, Non-Irrigated Arable Land, Broad-Leaved Forest.}
  \end{subfigure}
  \caption{\textbf{Learned perturbations (bottom right) appear varied in shape, intensity, and placement to augment semantic features in the original input in targeted ways.} The input image is on the left and resulting views are given on the top right. Perturbations (deltas) were generated using Viewmaker and then linearly scaled to the full image range for clear visualization with corresponding views.}
 \label{fig:view samples}
\end{figure*}

\section{Introduction}
\label{sec:intro}

Contrastive learning methods have demonstrated remarkable ability to learn high-quality representations without relying on labels, often achieving equivalent or higher classification accuracy than supervised approaches after pretraining on large unlabeled image datasets \cite{DBLP:journals/corr/abs-1805-01978, tamkin2022feature, DBLP:journals/corr/abs-2002-05709}.
These advances suggest the utility of contrastive learning in settings beyond natural images, including many impactful applications in the sciences, engineering, medicine, and beyond. However, a key barrier to mainstreaming contrastive learning applications is choosing the appropriate ``views'': the data corruptions that determine the contrastive learning process \cite{DBLP:journals/corr/abs-2005-10243, DBLP:journals/corr/abs-2002-05709}. It is challenging to develop effective views for new applications, as this process requires domain knowledge and trial and error for each setting.

Aerial satellites measure terabytes worth of multispectral image data in various forms every day. They use remote sensors to measure distinct wavelengths of light and stack outputs into $n$-channel images, which are used to analyze light beyond the human-visible spectrum. Such images can offer insights into natural phenomena like temperature variation that RGB data cannot. Robust and accurate self-supervised learning techniques for satellite images could aid advances in agricultural growth efficiency, understanding of climate change, tracking urban development, and environmental monitoring \cite{doi:10.1080/01431160701736471, NOH201252, PHINN20083413, Hanif_2019, 7535105}. In this paper, we investigate contrastive learning with multispectral satellite images.

The dominant views for contrastive learning on natural images were identified via extensive trial and error, and involve applying augmentations such as color jitter and horizontal flipping to RGB images \cite{NEURIPS2020_4c2e5eaa, DBLP:journals/corr/abs-2005-13149}. However, these RGB views do not transfer well to multispectral images since each channel has different numeral ranges and semantics, making it impossible to directly apply such transformations. Domain-agnostic generative Viewmaker networks \cite{DBLP:journals/corr/abs-2010-07432} propose to \textit{learn} such data transformations with a generative model trained adversarially with an encoder. However, Viewmaker networks have not yet been applied to a broader range of scientific data.

To address this gap, we evaluate Viewmaker networks on multispectral satellite image datasets. Furthermore, we introduce a generative network called Divmaker, which produces views optimized for diversity and does not require adversarial optimization. While Divmaker performs slightly worse than Viewmaker, our results confirm that both domain-agnostic view generation methods enable higher quality contrastive learning than with reasonable, hand-designed augmentations. We demonstrate this on three different large-scale multispectral satellite datasets and compare with an RGB satellite dataset for additional context.

\begin{figure*}
  \centering
  \begin{subfigure}{\linewidth}
    \includegraphics[width=\textwidth]{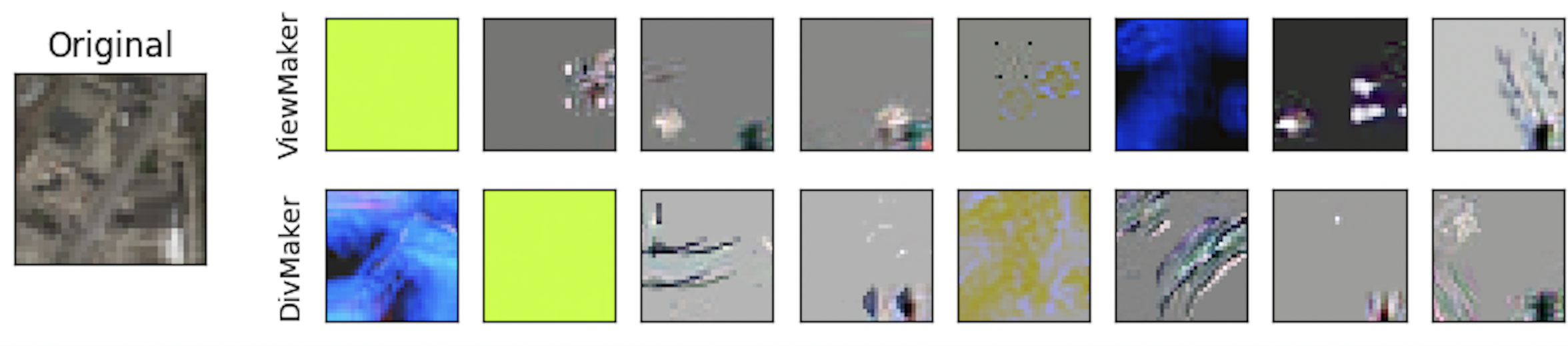}
    \caption{The input image is from NWPU-RESISC45 with the label Baseball Diamond.}
  \end{subfigure}
  \hfill
  \begin{subfigure}{\linewidth}
    \includegraphics[width=\textwidth]{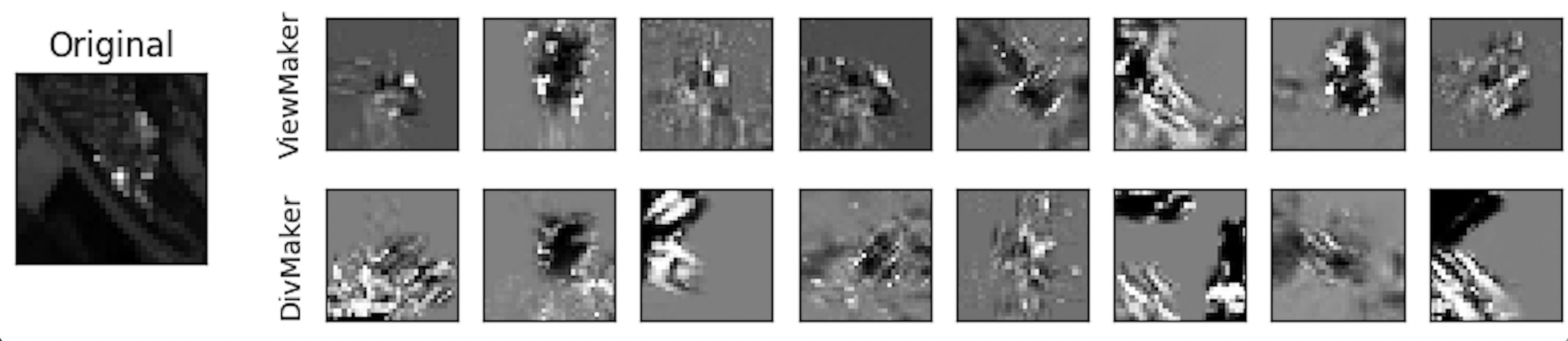}
    \caption{The input image is band 1 from EuroSAT with the label Highway.}
  \end{subfigure}
  \hfill
  \begin{subfigure}{\linewidth}
    \includegraphics[width=\textwidth]{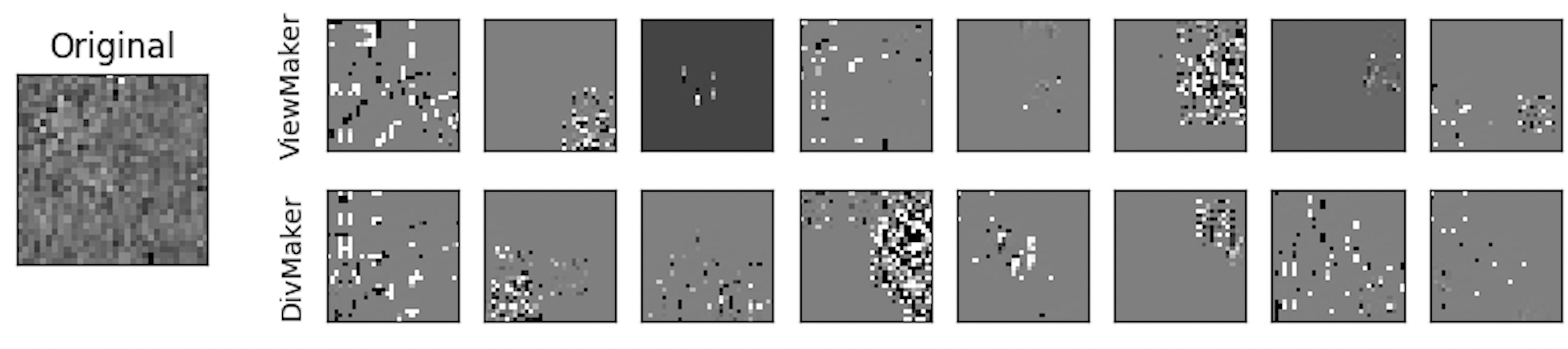}
    \caption{The input image is band 2 from So2Sat Sentinel-1 with the label Dense Trees.}
    \label{fig:so2sat sen1 adv div}
  \end{subfigure}
  \caption{\textbf{Divmaker manages to generate diverse and effective views without adversarial optimization.} We see this as perturbations from the Divmaker (bottom right) appear to act in similar ways to perturbations from the Viewmaker (upper right). The original input image is given on the left and perturbations (deltas) have been linearly scaled to the full image range for clear visualization.}
 \label{fig:adv div samples}
\end{figure*}

\section{Related Work}

\textbf{Contrastive learning}. Self-supervised learning enables learning representations from large, unlabeled datasets which can be used for downstream tasks like classification, object detection, semantic segmentation, or visual navigation. Contrastive learning methods have been shown to produce good representations by identifying transformed positive ``views'' of the same inputs \cite{DBLP:journals/corr/abs-1805-01978, DBLP:journals/corr/abs-2002-05709, DBLP:journals/corr/abs-2003-04297, DBLP:journals/corr/abs-2104-14294, DBLP:journals/corr/abs-2011-09157,
DBLP:journals/corr/abs-2006-07733}. Such approaches rely on good choices of data augmentations \cite{NEURIPS2020_4c2e5eaa, DBLP:journals/corr/abs-2005-13149} that yield representations discriminative with respect to the downstream tasks and yet general enough to be applied to new tasks. Such augmentations may not always be known a priori. This problem is especially pertinent in less common domains and modalities like multispectral satellite images, 3D images, tabular data, and voice recordings.

\textbf{Learning from satellite images.} Deep learning research on satellite imagery contends with a variety of factors including vast unlabeled datasets, spatial-temporal heterogeneity within classes, cloud interference, and texture and color discontinuities between image tiles \cite{https://doi.org/10.48550/arxiv.2211.08129}. Learning methods have been productively applied to tasks such as poverty mapping \cite{gram2019mapping}, local climate zone classification \cite{DBLP:journals/corr/abs-1912-12171}, water temperature prediction \cite{399146}, food safety analysis \cite{QIN2013157}, enhancing agricultural yield \cite{rs70404026}, everyday scene classification \cite{schmitt2019sen12ms, DBLP:journals/corr/abs-1709-00029, DBLP:journals/corr/ChengHL17}, and sustainable development monitoring \cite{yeh2021sustainbench}. In this paper, we investigate classification of land use, local climate zones, and everyday scenes.

\textbf{Contrastive learning for satellite images}. Multispectral satellite images often contain heterogeneous backgrounds with significant structural information and varying resolutions depending on the remote sensor's settings \cite{DBLP:journals/corr/abs-2005-01094, rs14163995}. Since these characteristics differ from those of standard RGB images (i.e. in ImageNet or CIFAR-10), traditional contrastive learning augmentations do not transfer well to multispectral images. Features like geographical distance between patches have been shown to produce useful views for datasets \cite{DBLP:journals/corr/abs-1805-02855}, although this requires having access to the coordinate location of each image or very-high resolution allowing the images to be divided into smaller patches during training \cite{rs14163995}. Another work splits input images into two views based on their channels, passes them as inputs to two different encoder networks, and uses the embeddings themselves as positive views for contrastive learning \cite{DBLP:journals/corr/abs-2011-09980}. Since each remote sensing application has different bands, the splitting process would need to be customized. There have been additional methods proposed like sharpness transformation  and random erasure \cite{DBLP:journals/corr/abs-2104-07070, rs15030827}, however many of these depend on dataset-specific properties (i.e. high resolution and number of bands) or have varying effectiveness across datasets. Due to the diversity of multispectral image formats, it is less feasible to exhaustively search for the most effective augmentation strategy as in \cite{DBLP:journals/corr/abs-2002-05709}, which isolated the best augmentation pairs on RGB images. Thus, we focus on general out-of-the-box view-generating methods that need little customization.

\textbf{Domain-agnostic machine learning.} Domain-specific self-supervised learning algorithms have enabled significant gains in fields such as natural language processing \cite{DBLP:journals/corr/abs-1810-04805, DBLP:journals/corr/abs-2003-10555}, computer vision \cite{DBLP:journals/corr/abs-2002-05709, DBLP:journals/corr/abs-2003-04297}, and speech processing \cite{DBLP:journals/corr/abs-2006-11477}. However, many other domains with rich, unlabeled datasets could also benefit from self-supervised approaches. Recent work responds to this interest in advancing domain-agnostic self-supervised learning with new benchmarks \cite{DBLP:journals/corr/abs-2111-12062, tamkin2022dabs} and learning algorithms \cite{DBLP:journals/corr/abs-2010-08887, DBLP:journals/corr/abs-2011-04419}. In this work, we apply and build on Viewmaker networks and demonstrate that such domain-agnostic learning approaches can be fruitfully applied in a range of different remote sensing applications.

\section{Methods}
In this section, we discuss Viewmaker and introduce a new variant called Divmaker, which optimizes for diverse view generation.

\subsection{Viewmaker}

The Viewmaker \cite{DBLP:journals/corr/abs-2010-07432} is a generative network trained to produce augmented images, or views, that are useful for contrastive learning. The Viewmaker $V$ takes an input image $X$ and gives a perturbation $V(X)$, which is added to the input to obtain the view $X + V(X)$. Adversarial training encourages perturbations to be complex and strong enough to necessitate encoding useful representations. Perturbations are constrained to an $l_1$ sphere (with size controlled by a distortion budget hyperparameter) around the input to maintain faithfulness to the original features. Lastly, the network injects random noise so perturbations differ from each other. Viewmaker-learned views have seen recent success, outperforming baseline augmentations on speech recordings and wearable sensor data and attaining comparable downstream accuracy on natural images. Note that the Viewmaker's adversarial training setup with the encoder requires a fully differentiable objective.

\subsection{Diversity Viewmaker (Divmaker)}

We introduce the Divmaker, a domain-agnostic view generation approach that does not require adversarial training and optimizes for diverse rather than challenging views. Like the Viewmaker, Divmaker offers a way to generate new views and uses the same contrastive loss.

Before formalizing the Divmaker view-generation loss, we introduce the following notation. Consider a temperature parameter $\tau$, an anchor input $x \in \mathcal{D}$ with embedding $z$, and associated views $x_i$ with embeddings $z_i$ for $1 \leq i \leq K$. We use the cosine similarity measure on embeddings:
\begin{equation}
\text{sim}(z,z')=\frac{z^T z'}{||z||||z'||}
\end{equation}
And define:
\begin{equation}
h(z, z') = \exp\left(\frac{\text{sim}(z,z')}{\tau}\right)
\end{equation} 
Then the Divmaker loss is:
\begin{equation}
\mathcal{L} = \textbf{E}_{x \sim \mathcal{D}} \left(- \sum_{k=1}^K \log{\frac{h(z_k, z)}{h(z_k,z) + \sum_{l \neq k} h(z_k, z_l)}}\right)
\end{equation}

The intuition is to optimize diversity by maximizing cosine similarity between the original input and its generated views while minimizing similarity between two views ($K = 2$) for the same input. This is in contrast to the Viewmaker, which tries to create challenging views via an adversarial loss without explicitly optimizing for the diversity of the views. This diversity objective was previously used for self-supervised anomaly detection \cite{DBLP:journals/corr/abs-2103-16440}, where the views learned were a finite set of learned masks. Instead, we generate dynamic and input-conditioned views with a stochastic neural network, as Viewmaker does. The Divmaker loss is also closely related to the Triplet loss \cite{NIPS2005_a7f592ce}, which has been widely used in other applications and shares parallels with standard contrastive learning losses \cite{NEURIPS2020_d89a66c7}.

Like the Viewmaker, the Divmaker network outputs a bounded perturbation, which is added to the input to produce a view that can be used for contrastive learning. The strength of  Divmaker-generated views is controlled by the distortion budget, which specifies the magnitude of Divmaker's perturbation output as done in the Viewmaker. Training with diverse views could help capture a wider range of augmentations encountered in practice and enable the encoder to learn useful representations earlier in training. Furthermore, by separating the Divmaker and encoder objectives, we eliminate the differentiable restriction on the encoder, allowing Divmaker to work with state-of-the-art non-differentiable contrastive learning methods \cite{DBLP:journals/corr/abs-2104-14294, DBLP:journals/corr/abs-2006-09882}. Finally, while we did not experience training instabilities with Viewmaker, Divmaker's avoidance of adversarial training may enable more stable training for larger models \cite{DBLP:journals/corr/KodaliAHK17}. 

\subsection{Datasets}

\begin{figure*}
     \centering
     \begin{subfigure}{0.45\textwidth}
        \centering
        \includegraphics[width=\textwidth]{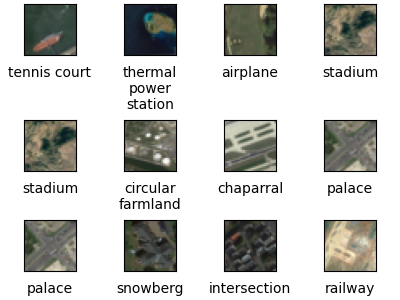}
         \caption{Twelve samples from the satellite NWPU-RESISC45 dataset.}
         \label{fig:single satellite}
     \end{subfigure}
     \hfill
     \begin{subfigure}{0.45\textwidth}
        \centering
        \includegraphics[width=\textwidth]{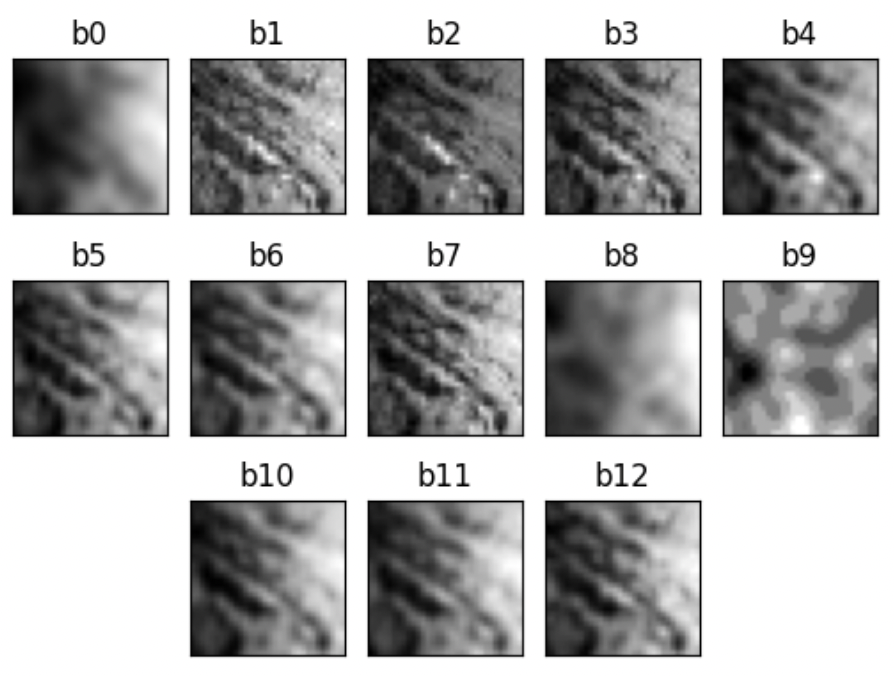}
        \caption{A single EuroSat sample with the label Sea and Lake.}
        \label{fig:single eurosat}
     \end{subfigure}
     \begin{subfigure}{0.45\textwidth}
        \centering
        \includegraphics[width=\textwidth]{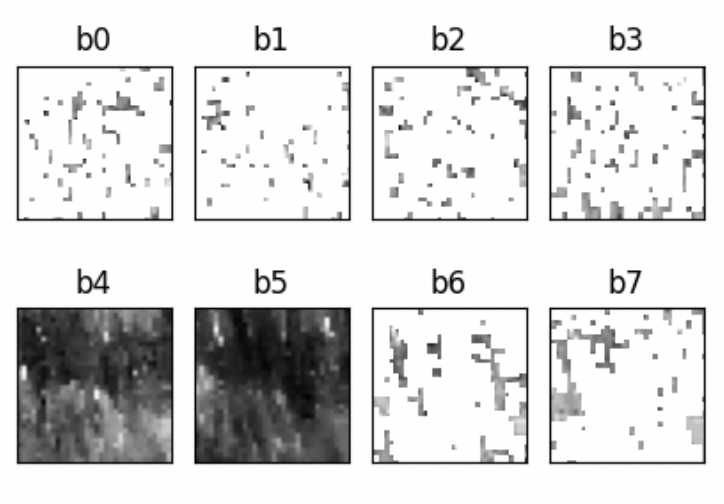}
        \caption{A single So2Sat Sentinel-1 sample with the label Open High-Rise.}
        \label{fig:single so2sat s1}
     \end{subfigure}
     \hfill
     \begin{subfigure}{0.45\textwidth}
        \centering
        \includegraphics[width=\textwidth]{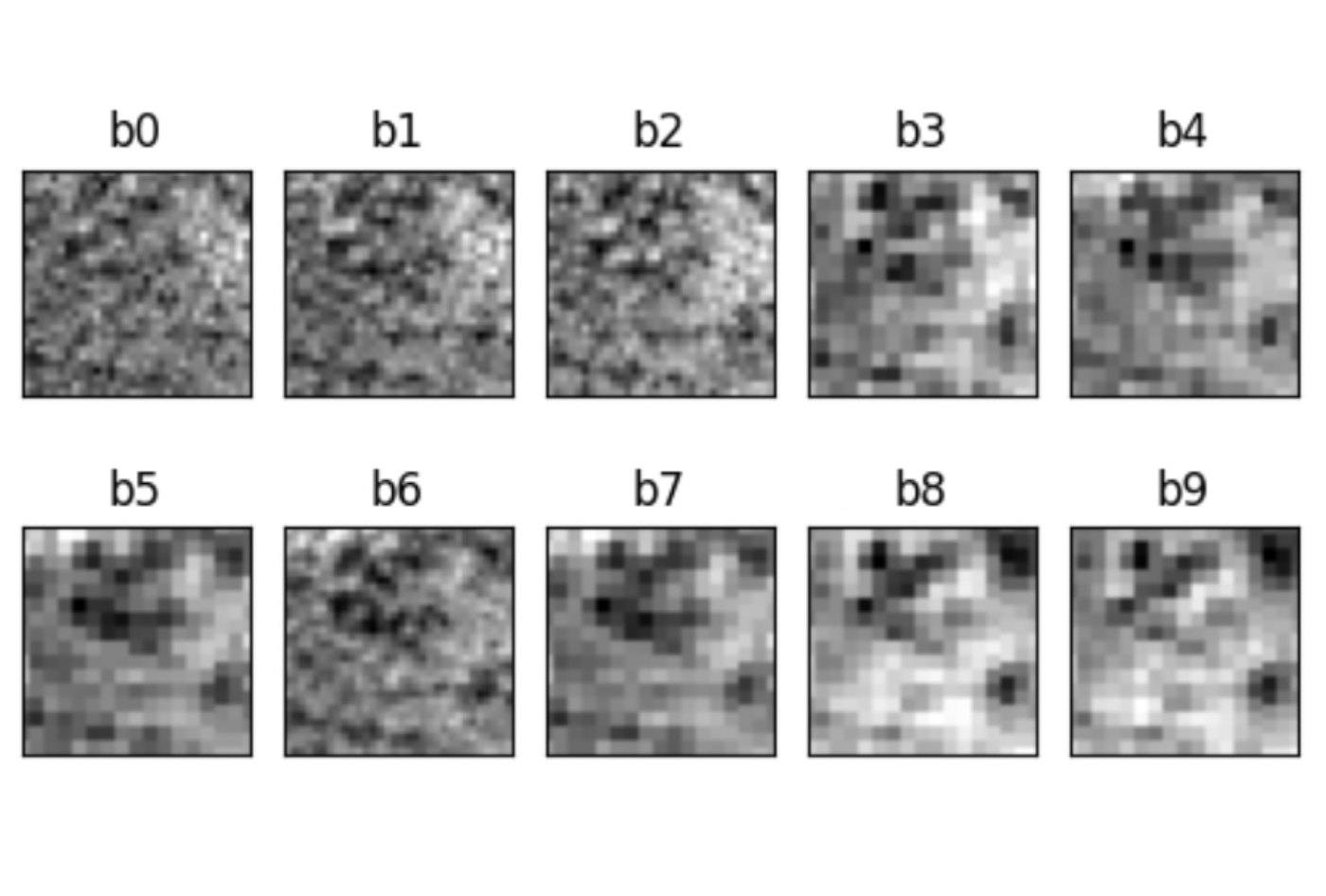}
        \caption{A single So2Sat Sentinel-2 sample with the label Dense Trees.}
        \label{fig:single so2sat s2}
     \end{subfigure}
     \begin{subfigure}{0.45\textwidth}
        \centering
        \includegraphics[width=\textwidth]{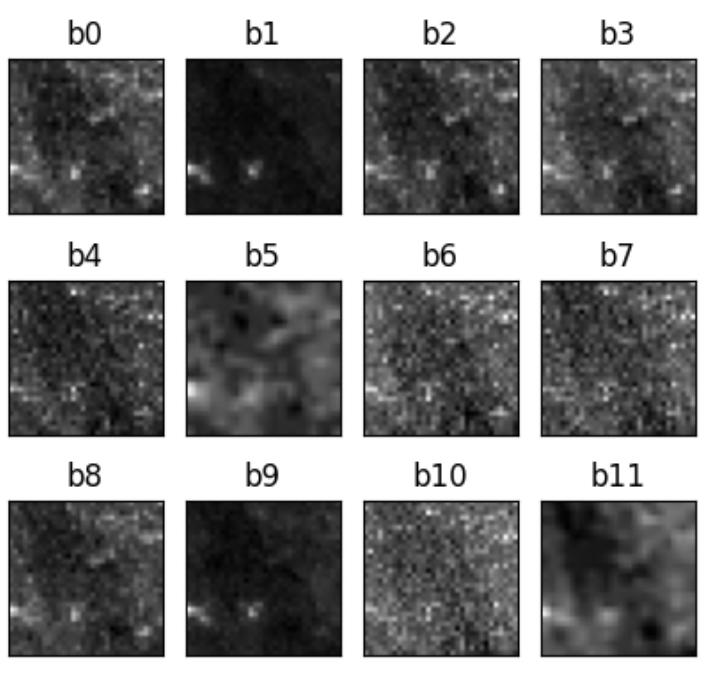}
        \caption{A single BigEarthNet sample with the multi-label: Sea and Ocean, Water Bodies.}
        \label{fig:single bigearthnet}
     \end{subfigure}
     \caption{\textbf{Multispectral satellite images have very different characteristics from RGB images, and thus require different strategies for creating views.} We display randomly selected images from each dataset considered. All channels are shown for multispectral images with band number prefaced with "b".}
     \label{fig:single_data}
\end{figure*}

\begin{table*}
\centering
\begin{tabular}{lccccl}\toprule
 Dataset  & Metric & Basic Expert & Expert & Viewmaker & Divmaker \\\midrule
 NWPU-RESISC45 & Accuracy & N/A & \textbf{76.07} & 67.61 & 71.72 \\ 
 EuroSAT & Accuracy & 90.93 & 90.68 & \textbf{96.4} & 95.67 \\
 So2Sat Sentinel-1 & Accuracy & 31.12 & 29.33 & \textbf{38.25} & 36.39 \\
 So2Sat Sentinel-2 & Accuracy & 51.6 & 51.54 & \textbf{60.08} & 59.67 \\
 BigEarthNet & F1 Score & 4.03 & 4.21 & \textbf{12.99} & 10.83 \\\bottomrule
\end{tabular}
\caption{\textbf{Domain-agnostic methods outperform domain-specific approaches on satellite datasets.} We measure performance as linear classification accuracy over pre-trained representations except on BigEarthNet, for which we use F1 score of multi-classification accuracy. For multispectral datasets, random cropping makes up the basic expert augmentations with horizontal flipping added for expert views. For NWPU-RESISC45, we use standard expert RGB augmentations. Results are averaged over four seeds with tuned distortion budgets.}
\label{table:accuracies}
\end{table*}

\begin{figure*}
\centering
   \includegraphics[width=0.97\linewidth]{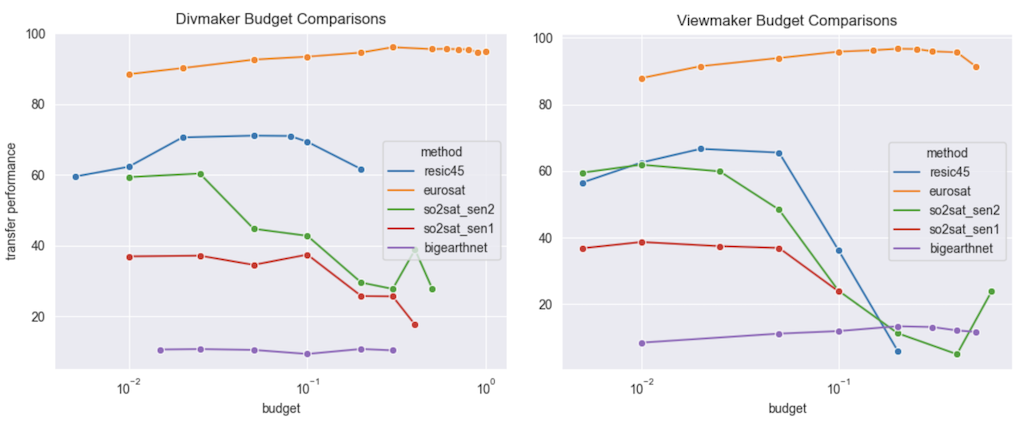}
\caption{Downstream task performance can change drastically with the distortion budget, but a wide range of settings enable good performance.}
\label{fig:budget_graphs}
\end{figure*}

We apply Viewmaker and Divmaker to four large-scale satellite datasets. Examples from each dataset are given in Figure \ref{fig:single_data}. Note that each band is shown separately for multispectral images.
\smallbreak
\textbf{EuroSAT} is a dataset of 27,000 images consisting of low cloud-cover satellite Sentinel-2 images from 34 European countries. Images are labeled with one of 10 classes describing land use such as Sea and Lake, Industrial, and Pasture with 2,000 to 3,000 images per class. Each image includes 13 spectral bands in the visible, near infrared, and short wave infrared part of the light spectrum \cite{DBLP:journals/corr/abs-1709-00029}.
\smallbreak
\textbf{So2Sat LCZ42 Sentinel-1 and Sentinel-2} is a benchmark dataset composed of 400,673 low cloud-cover images collected by Sentinel-1 and Sentinel-2 satellites over 42 large cities and 10 smaller areas spanning six continents. Images are labeled with one of 17 classes in the Local Climate Zones (LCZ) classification scheme, which are based on climate-relevant surface properties such as structure, surface cover, and anthropogenic parameters. Some examples include Open High-Rise, Dense Trees, Heavy Industry, and Water \cite{DBLP:journals/corr/abs-1912-12171}. Sentinel-1 data contains 8 real-valued bands Sentinel-2 data contains 10 real-valued bands.
\smallbreak
\textbf{BigEarthNet} is a multi-label dataset made up of 590,326 Sentinel-2 image patches collected from 10 European countries. Each image is labeled with a subset of 43 land-cover classes such as Pastures, Water Bodies, Agro-Forestry Areas, and Green Urban Areas \cite{DBLP:journals/corr/abs-1902-06148}. Note that, in the BigEarthNet dataset, two bands measure $20 \times 20$ images, six bands measure $60 \times 60$ images, and four bands measure $120 \times 120$ images. To standardize the full size, we resize all bands to $120 \times 120$ resolution and then stack the bands.
\smallbreak
\textbf{NWPU-RESISC45} is a dataset of 31,500 RGB satellite images sourced from Google Earth. Images are labeled with one of 45 classes such as Tennis Court, Thermal Power Station, Airplane, Stadium, Circular Farmland, Chaparral, Palace, Snowberg, Intersection, and Railway. The dataset includes 700 samples for each scene class with large variations in translation, spatial resolution, viewpoint, object pose, illumination, background, and occlusion. This dataset is challenging due to large variance within-class and high inter-class similarities \cite{DBLP:journals/corr/ChengHL17}. Since the NWPU-RESISC45 dataset consists of RGB satellite images, we use it as a control to compare Viewmaker performance when we have more domain knowledge and better expert views. 

\section{Experiments}

In this section, we explore whether Viewmaker and Divmaker can outperform domain-specific methods by learning to generate appropriate views on four well-known satellite datasets.

\subsection{Experimental Details}

We first train the encoder and view-generating networks simultaneously by pretraining on a dataset. Then we evaluate the quality of the learned representations using the widely used linear transfer protocol \cite{DBLP:journals/corr/abs-1807-03748, DBLP:journals/corr/abs-1805-08974, DBLP:journals/corr/abs-2002-05709}. We evaluate on the RGB NWPU-RESISC45 dataset as a baseline to confirm that well-researched expert views from the RGB image domain can give greater performance gains \cite{DBLP:journals/corr/abs-2002-05709}. We use random cropping as the basic expert transformation and combine this with horizontal flipping for expert views for the other datasets. While there exist more complex augmentation methods \cite{rs15030827, DBLP:journals/corr/abs-1805-02855, DBLP:journals/corr/abs-2104-07070}, we choose these because they are generalizable to any large-scale satellite dataset and hence useful benchmarks in the domain-agnostic setting.

For pretraining and downstream task training, we use a learning rate of $0.005$ and the same encoder and viewmaker architectures, temperature parameter, batch size, and other parameters as \cite{DBLP:journals/corr/abs-2010-07432}. We find normalizing multispectral images before passing through the Viewmaker and Divmaker networks to be useful for all datasets except NWPU-RESISC45, which has low standard deviation across channels before normalization. We also clamp all pixels in generated views symmetrically between $-1$ and $1$. For the Divmaker loss, we experimented with $K = 2$ and $K = 3$, finding that $K = 3$ gave minimal performance gains for greater computational cost. Hence, we used $K = 2$ in all reported experiments throughout the paper. All implementation details are available in our open source code at \url{https://github.com/jbayrooti/divmaker}.

\subsection{Results and Interpretation}

In Table \ref{table:accuracies}, we report the best performance for every method on each dataset. For the NWPU-RESISC45 dataset, learning from expert RGB transformations results in the highest linear classification accuracy over learned Viewmakers. This is not surprising, considering the abundance of research into these RGB transformations \cite{DBLP:journals/corr/abs-2002-05709, NEURIPS2020_4c2e5eaa}. For the multispectral datasets, we find much stronger gains from Viewmaker and Divmaker methods, compared to the handcrafted augmentations: horizontal flipping and cropping. This demonstrates the utility of domain-agnostic methods like Viewmaker and Divmaker when working with less popular data forms.

Divmaker performs better than Viewmaker on the RGB dataset and nearly as well on the multispectral datasets, indicating that a broader range of objectives for generative views can enable success on multispectral data. We provide illustrations of views and perturbations on NWPU-RESISC45 in Figure \ref{fig:view samples} and a comparison of perturbations produced by Viewmaker and Divmaker in Figure \ref{fig:adv div samples}. Related work demonstrates that, in some cases, the Viewmaker network can identify and alter semantic features in an input to aid learning \cite{tamkin2022feature}. Although it is hard to pinpoint exact interpretations of the perturbations, our work corroborates this as perturbations appear correlated across channels and sometimes with simple image features.

We also compare downstream classification performance for Viewmaker and Divmaker with different distortion budgets in Figure \ref{fig:budget_graphs}. This demonstrates that, while the budget should be tuned for optimal performance, a wide range of budgets allow for good performance. These results also confirm that Divmaker can achieve comparable performance with Viewmaker for slightly higher budgets. This is expected since Divmaker does not use adversarial training.

\section{Conclusion}

In this paper, we examine whether domain-agnostic approaches to contrastive learning can scale to an important scientific domain: multispectral satellite images. Our experiments show that domain-agnostic methods can outperform existing domain-specific contrastive learning methods using out-of-the-box baseline views. This insight is important considering the utility of multispectral satellite images, and suggests that domain-agnostic self-supervised methods may enjoy success across a wider array of scientific applications. Additionally, we demonstrate another successful strategy for domain-agnostic view learning with the Divmaker, which avoids adversarial optimization. Future directions of research include comparing against more sophisticated domain-specific view-generation methods, analyzing further downstream differences between the Viewmaker and Divmaker, and considering additional multispectral applications like water temperature prediction \cite{399146} and sustainable development monitoring \cite{yeh2021sustainbench}.

{\small
\bibliographystyle{ieee_fullname}
\bibliography{egbib}
}

\end{document}